\documentclass{article}
\usepackage[main,final]{neurips_2026}
\makeatletter
\renewcommand{\@notice}{}
\makeatother
\usepackage[utf8]{inputenc}
\usepackage[T1]{fontenc}
\usepackage{microtype}
\usepackage{graphicx}
\usepackage{subcaption}
\usepackage{booktabs}
\usepackage{hyperref}

\usepackage{xcolor}
\usepackage{colortbl}

\definecolor{vhband}{RGB}{226,239,231}
\definecolor{alfredband}{RGB}{232,228,241}
\definecolor{basicband}{RGB}{229,236,247}
\definecolor{complexband}{RGB}{253,237,211}

\usepackage{adjustbox}
\usepackage{amsmath}
\usepackage{amssymb}
\usepackage{amsthm}

\usepackage{tikz}
\theoremstyle{remark}

\theoremstyle{plain}
\newtheorem{theorem}{Theorem}

\newtheorem{proposition}[theorem]{Proposition}

\usepackage{pgfplots}
\pgfplotsset{compat=1.18} 
\usepgfplotslibrary{groupplots} 

\usepackage{makecell}
\usepackage{tabularx}
\usepackage{array}

\newcolumntype{L}{>{\raggedright\arraybackslash}X}
\newcommand{\ms}[2]{%
    \ensuremath{#1{\scriptstyle\,\pm\,#2}}%
}

\title{QuantWAMs: Calibrating at the Right Granularity for World Action Models}

\author{%
  Jiacheng Zhou\textsuperscript{1} \quad
  Jinfan Lv\textsuperscript{1} \quad
  Ruixuan Li\textsuperscript{1} \quad
  Yan Wang\textsuperscript{3} \\
  \textbf{Longtai Zhang\textsuperscript{1}} \quad
  \textbf{Wenqiang Zhang\textsuperscript{1,2}} \quad
  \textbf{Lizhe Qi\textsuperscript{1}} \\[0.5ex]
  \normalfont
  \textsuperscript{1}College of Intelligent Robotics and Advanced Manufacturing, Fudan University\\
  \textsuperscript{2}Shanghai Key Lab of Intelligent Information Processing,\\
  College of Computer Science and Artificial Intelligence, Fudan University\\
  \textsuperscript{3}School of Data Science and Engineering, East China Normal University\\[0.5ex]\\
  \textnormal{Project page: \url{https://quantwams.github.io}}
}

\begin{document}

\maketitle

\begin{abstract}
World Action Models (WAMs) jointly predict future observations and actions,
but their iterative denoising and closed-loop execution make efficient deployment costly.
Existing post-training quantization (PTQ) methods are poorly suited to WAMs because they rely on open-loop objectives,
homogeneous model assumptions, and calibration distributions that do not reflect deployment.
We present QuantWAMs, a PTQ framework that aligns quantization decisions with the calibration context defined by model structure,
rollout distribution, and task objective. QuantWAMs introduces three strategies:
shared-basis outlier calibration, which pools activation evidence only across coordinate-compatible modules;
co-training-objective saliency, which computes empirical-Fisher scores from the joint video--action gradient and assigns weight precision at a calibration-stable layer granularity;
and fixed-intervention rollout auditing, which revises denoising-step protection schedules using reachable closed-loop states without changing the precision budget.
We evaluate QuantWAMs on Fast-WAM and LingBot-VA across RoboTwin 2.0, LIBERO, and real-robot manipulation with an AgiBot G2.
Under a W4A4-dominant setting, the reported simulation means differ from FP16 by 0.2--0.7 percentage points.
Real-robot trials further establish deployment feasibility on three manipulation tasks.
For the targeted video and action blocks, QuantWAMs reduces peak weight-and-activation memory to about 29\% of FP16 and provides 1.4--1.6$\times$ block-level speedups.
\end{abstract}

\begin{figure}[h]
\label{fig:outlier_evolution}
\centering
\includegraphics[width=\textwidth]{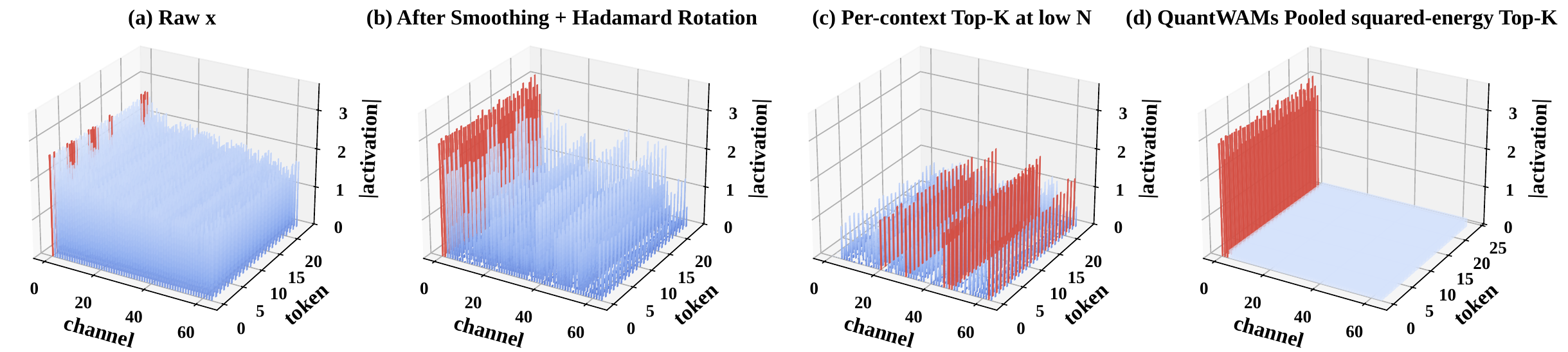} 
\caption{
Outlier evolution on self\_attn.o.
(a) Raw activation shows persistent per-channel outliers.
(b) Joint Smoothing + Hadamard rotation disperses magnitude across channels.
(c) Per-context Top-K at low N samples noise yields unstable masks and lower recovered energy.
(d) QuantWAMs preserves the pooled-energy Top-\(K\) channels in FP16 and quantizes others to A4.
}
    \vspace{-1em}
\end{figure}

\section{Introduction}
\label{sec:intro}
World Action Models integrate a video diffusion world model with a diffusion action expert
to jointly predict future observations and the actions that realize them, a promising
paradigm for general purpose robotic manipulation that is costly to deploy
\cite{yuan2026fast,li2026causal,ye2026world,bi2025motus,cen2025worldvla,liang2025video}.
Each control
cycle requires iterative denoising, and proceeds in closed loop, where every generated
action changes the observations used by later predictions
\cite{chi2025diffusion,peebles2023scalable,li2026causal,ye2026world}.
Post-training quantization is
therefore a natural route to real-robot deployment.

Yet most PTQ methods rely on two assumptions that fail for WAMs. They optimize open loop
objectives such as perplexity or single image reconstruction, where each forward pass is
scored independently, and they assume a homogeneous transformer
\cite{frantar2022gptq,xiao2023smoothquant,lin2025awq,chen2025q,he2023ptqd,so2023temporal,shang2023post,li2023q}.
WAMs instead propagate
early quantization error into later states through closed loop interaction
\cite{ross2011reduction}, and distribute
computation across coupled pathways, either dual stream Mixture of Transformers or a shared
diffusion backbone
\cite{yuan2026fast,li2026causal,ye2026world,bi2025motus}.
A static single stream proxy may mis-allocate precision even
when its calibration loss remains low.

During our experiments to transfer quantization algorithms from classic large models and DiT models to WAMs,
we found that every PTQ decision is a finite-sample estimate fixed before deployment under a
\emph{calibration context}: the scope $\mathcal{G}$ over which evidence is pooled, the state
distribution $\widehat{\mathcal{D}}$ on which it is measured, and the objective
$\mathcal{L}$ under which it is scored. A decision is useful only if all three remain
appropriate at deployment. A \emph{structural} mismatch pools evidence across members without a
common coordinate system, or splits it until sampling noise dominates. A
\emph{distributional} mismatch profiles sensitivity on states unreachable in closed loop,
producing phantom denoising-step peaks that misdirect the precision budget. An
\emph{objective} mismatch scores a single stream, or fuses streams after squaring,
discarding a coordinatewise interaction present in the joint video--action gradient. The three are not separate
heuristics but one requirement applied along structural, distributional, and objective axes.
We screen each decision at the granularity supported by the available calibration data.

Based on this principle, we propose QuantWAMs, a PTQ algorithm for WAMs.
We evaluate it on RoboTwin 2.0\cite{chen2025robotwin}, LIBERO\cite{liu2023libero} and real-robot manipulation with
AgiBot G2 using two released WAM implementations
\cite{yuan2026fast,li2026causal}.
The same calibration procedure applies to both implementations with
architecture-specific grouping rules. Under the W4A4-dominant setting, its
reported simulation means differ from FP16 by 0.2--0.7 percentage
points.
Our contributions are threefold:
\begin{itemize}
\item We formulate post-training quantization for WAMs around the structural, distributional, and objective contexts that determine a deployment-time precision decision.
\item We derive when activation evidence may be pooled across modules or modes and characterize the sample-heterogeneity crossover governing Top-$K$ mask recovery.
\item We jointly score video--action gradients at a calibration-stable layer granularity and use fixed-intervention replay on real rollouts to revise denoising-step schedules at a fixed precision budget.
\end{itemize}

\section{Related Work}

\paragraph{Post-training quantization.}
PTQ for large models follows four main directions.
Weight-only methods \cite{frantar2022gptq} use second-order error compensation,
while \cite{zhao2024atom} protects salient channels through activation-aware scaling.
Weight-activation methods include SmoothQuant\cite{xiao2023smoothquant},
which migrates activation outliers into weights, and \cite{zhao2024vidit},
which calibrates timestep-varying distributions in diffusion transformers.
For activation offloading, we retain Atom's square-sum channel statistic
and mixed-precision execution primitive\cite{zhao2024atom}; our contribution is to determine
the calibration contexts over which this statistic may be pooled.
Weight-precision allocation is treated separately through a
joint-objective empirical-Fisher score.

\paragraph{Quantizing robot policies.}
Recent work compresses vision-language-action models for embodied
deployment, establishing that low-bit manipulation policies are
feasible\cite{zhang2026quantvla,xu2026qvla,park2025saliency}.
These methods target a single vision-language backbone with an action head and
validate on open-loop surrogates such as action reconstruction error. Neither which modules
may share calibration evidence nor how error compounds across a control loop arises there.

\paragraph{World Action Models.}
World Action Models introduce two properties absent from these settings. Deployment is
closed loop, so early quantization error shifts the states encountered later and fidelity on
a fixed input distribution need not predict task success
\cite{yuan2026fast,li2026causal,ye2026world,ross2011reduction}.
Multi-stream and shared-backbone
architectures couple heterogeneous paths
\cite{yuan2026fast,li2026causal,ye2026world,bi2025motus},
so whether two modules may share a calibration
statistic is fixed by the deployment distribution rather than by convenience.

\begin{figure}[t]
\centering
\includegraphics[width=\textwidth]{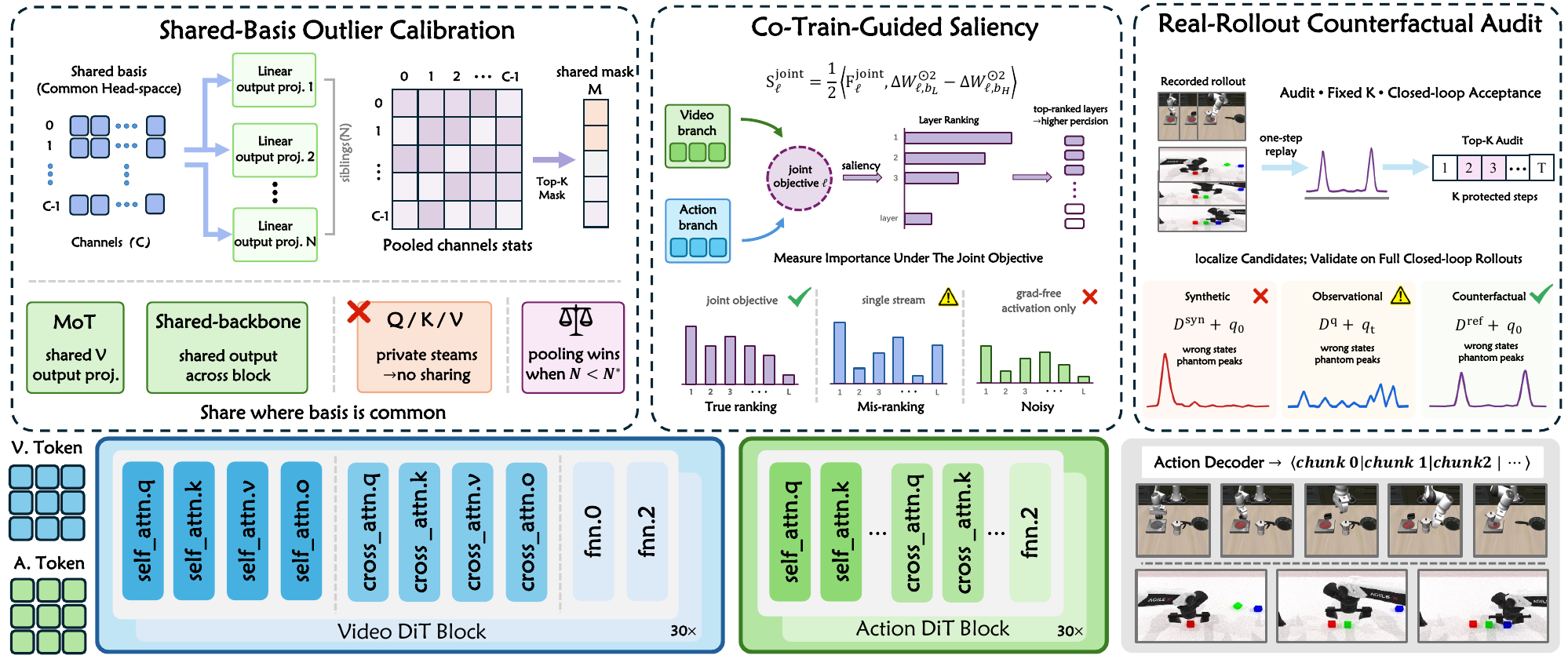} 
\caption{
Overview of QuantWAMs. Shared-basis calibration pools activation statistics only across coordinate-compatible modules;
co-training-guided saliency allocates precision using joint video–action gradients;
and fixed-intervention rollout auditing revises denoising-step schedules for coupled video and action DiT blocks.
}
\label{fig2}
\vspace{-1em}
\end{figure}

\section{Method}
\label{sec:method}
Key PTQ decisions, including outlier preservation, layerwise weight-bit allocation,
and timestep protection, constitute finite-sample estimates established prior to deployment.

\subsection{Shared-Basis Outlier Calibration}
\label{sec:sbc}

Outlier-aware activation quantization preserves
$K=\lfloor\rho d\rfloor$ input channels at high precision. We retain
Atom's mixed-precision channel-offloading primitive and study
only how its static mask should be shared. A context $i$ specifies a module,
execution mode, and depth. Let $A_i^{(n)}\in\mathbb R^{T_{i,n}\times d}$
be its activation on trajectory $n$ before the offline transforms. The
runtime quantizer receives
\begin{equation}
\widetilde A_i^{(n)}
=
\left(A_i^{(n)}R\right)S_i^{-1},
\label{eq:runtime-transform}
\end{equation}
where the Hadamard rotation $R$ is common within an admissible group and
$S_i$ is a context-specific diagonal smoothing matrix. All statistics below
are collected from this transformed input:
\begin{equation}
\begin{gathered}
z_i^{(n)}(c) = \frac{1}{T_{i,n}} \sum_{t=1}^{T_{i,n}} \left(\widetilde A_i^{(n)}\right)_{t,c}^{2},
\qquad
e_i(c) = \mathbb E[z_i^{(n)}(c)],
\\[1mm]
\widehat e_i(c) = \frac{1}{N_i}\sum_{n=1}^{N_i}z_i^{(n)}(c),
\qquad
e_g(c) = \sum_{i\in g}\pi_i e_i(c),
\\[1mm]
\widehat e_g(c) = \sum_{i\in g}\pi_i\widehat e_i(c).
\end{gathered}
\label{eq:sbc-statistics}
\end{equation}

Here $\sum_i\pi_i=1$, and $\pi_i$ encodes the intended deployment exposure
rather than token count. Under an equal-cost energy surrogate, the
shared-constrained oracle and its estimator are
\begin{equation}
\begin{aligned}
\Omega_g^\star
&=
\arg\max_{|\Omega|=K}
\sum_{i\in g}\pi_i\sum_{c\in\Omega}e_i(c),
\\
&=
\operatorname{TopK}_{K}(e_g),
\\
\widehat\Omega_g
&=
\operatorname{TopK}_{K}(\widehat e_g).
\end{aligned}
\label{eq:sbc-mask}
\end{equation}
This optimality is restricted to the stated surrogate and does not imply
optimal closed-loop performance.

\paragraph{Coordinate admissibility.}
A literal index can be shared only when it denotes the same quantizer-input
coordinate in every member.
Let a common representation
$Z\in\mathbb R^{n\times d}$ in a fixed ordered basis and
$A_i=P_iZ$, where $P_i$ acts only on token rows. For a column selector
$E_\Omega$ and the restricted diagonal matrix $S_{i,\Omega}$,
\begin{equation}
\widetilde A_iE_\Omega
=
P_i(ZR)E_\Omega S_{i,\Omega}^{-1}.
\label{eq:shared-coordinate}
\end{equation}
Thus the common rotation fixes one ordered basis throughout the group, while
the context-specific diagonal scaling changes magnitudes but not coordinate
identity. A context-specific dense rotation would invalidate literal
index sharing unless an explicit map to a common basis were supplied.
In Fast-WAM, the shared value pathway aligns paired output projections, whereas
expert-private residual streams do not license cross-expert Q/K/V sharing.
In the released shared-backbone LingBot-VA, modes reuse the input columns of
the same physical Linear, making cross-mode sharing admissible when
post-fusion column transforms are shared or coordinate-preserving. None of
these statements implies distributional equality. Pooling across depth is a
separate exchangeability assumption.

For a coordinate-admissible group, model each channel statistic as
\begin{equation}
\begin{aligned}
z_i^{(n)}(c)
&=
\mu_g(c)+\tau_c\eta_i(c)+\sigma_c\xi_i^{(n)}(c),
\\
e_i(c)
&=
\mu_g(c)+\tau_c\eta_i(c),
\end{aligned}
\label{eq:random-effects}
\end{equation}
where $\tau_c^2$ is cross-member heterogeneity and $\sigma_c^2$ is
within-member sampling variation. Under the balanced working model with
$m=|g|$, equal weights, $N$ independent trajectory units per member, and
independent sampling errors across members, the expected member-fidelity
risks are
\begin{equation}
\begin{aligned}
R_{\mathrm{ind},c}
&:=
\sum_{i\in g}\pi_i
\mathbb E\!\left[
\bigl(\widehat e_i(c)-e_i(c)\bigr)^2
\right]
=
\frac{\sigma_c^2}{N},
\\
R_{\mathrm{pool},c}
&:=
\sum_{i\in g}\pi_i
\mathbb E\!\left[
\bigl(\widehat e_g(c)-e_i(c)\bigr)^2
\right],
\\
&=
\frac{m-1}{m}\tau_c^2+\frac{\sigma_c^2}{mN}.
\end{aligned}
\label{eq:pooling-risks}
\end{equation}
\begin{proposition}[Pooling crossover]
\label{prop:crossover}
For $\tau_c^2>0$, pooling lowers risk exactly when
\begin{equation}
N<N_c^\star=\frac{\sigma_c^2}{\tau_c^2},
\qquad
N_{\mathrm{eff},c}
=
\frac{mN}{1+(m-1)N\tau_c^2/\sigma_c^2}.
\label{eq:pooling-crossover}
\end{equation}
If $\tau_c^2=0$, pooling dominates for every finite $N$.
\end{proposition}

Lower energy risk need not recover a member's preferred mask. Let
$\Omega_i^\star=\operatorname{TopK}_K(e_i)$,
$\Delta_i=e_{i,(K)}-e_{i,(K+1)}$, and
$\beta_i=\|e_g-e_i\|_\infty$. If $\beta_i<\Delta_i/2$ and the centered
pooled error is channel-wise sub-Gaussian with proxy $s_g^2$, then
\begin{equation}
\Pr\!\left[\widehat\Omega_g\neq\Omega_i^\star\right]
\leq
2d\exp\!\left[
-\frac{(\Delta_i/2-\beta_i)^2}{2s_g^2}
\right].
\label{eq:mask-recovery-main}
\end{equation}

The architecture first supplies candidate groups. For each benchmark, the
same fixed set of 32 training trajectories is used to estimate the group
statistics, fit the mask, and calibrate the remaining quantizer parameters.
We form a rank window $\widehat{\mathcal H}_g$ around $K$ and compute
\begin{equation}
\begin{aligned}
\widehat{\mathcal H}_g
&=
\left\{
c:
\left|
\operatorname{rank}_{\downarrow}
\bigl(\widehat e_g(c)\bigr)-K
\right|
\leq h
\right\},
\\
\widehat N_g^\star
&=
\operatorname*{median}_{c\in\widehat{\mathcal H}_g}
\frac{\widehat\sigma_c^2}
{\max(\widehat\tau_c^2,\varepsilon)}.
\end{aligned}
\label{eq:estimated-crossover}
\end{equation}
We pool only coordinate-admissible groups satisfying
$N_{\mathrm{cal}}<\widehat N_g^\star$ without degrading bootstrap mask
stability. The bootstrap resamples complete trajectories and preserves the
paired observations across contexts; it is a stability analysis of the same
32-trajectory calibration set, not a second fitting split. The final mask is
$\operatorname{TopK}_K(\widehat e_g)$ on all 32 trajectories.
Proofs, estimators, the covariance-aware bootstrap, and architecture-specific
grouping rules appear in Appendix A.

\subsection{The Weight Axis: Co-Training-Objective Saliency}
\label{subsec:cotrain}

The preceding section determines where calibration evidence may be pooled;
here the same principle determines the granularity of weight-precision
decisions. For a Linear $y_L=W_Lx_L$, we evaluate perturbations under the
video--action co-training objective used by the pretrained WAM,
\begin{equation}
\ell_{\mathrm{co}}
=
\lambda_{\mathrm{v}}\ell_{\mathrm{v}}
+
\lambda_{\mathrm{a}}\ell_{\mathrm{a}}.
\label{eq:cotrain-objective}
\end{equation}
For each of the 32 paired calibration trajectories, let
$g_{m,L}=\nabla_{y_L}\ell_m$, $m\in\{\mathrm{v},\mathrm{a}\}$, and define
\begin{equation}
\Sigma_L=\mathbb{E}[x_Lx_L^\top],
\qquad
G_L^{m}=\mathbb{E}[g_{m,L}g_{m,L}^{\top}].
\label{eq:weight-factors}
\end{equation}
Computing the empirical Fisher after combining the two losses gives
\begin{equation}
\begin{aligned}
G_L^{\mathrm{joint}}
&=
\mathbb{E}\!\left[
(\lambda_{\mathrm v}g_{\mathrm v,L}
+\lambda_{\mathrm a}g_{\mathrm a,L})
(\lambda_{\mathrm v}g_{\mathrm v,L}
+\lambda_{\mathrm a}g_{\mathrm a,L})^\top
\right]
\\
&=
\lambda_{\mathrm v}^{2}G_L^{\mathrm v}
+\lambda_{\mathrm a}^{2}G_L^{\mathrm a}
+\lambda_{\mathrm v}\lambda_{\mathrm a}
\left(\Xi_L+\Xi_L^\top\right),
\\
G_L^{\mathrm{fusion}}
&=
\lambda_{\mathrm v}^{2}G_L^{\mathrm v}
+\lambda_{\mathrm a}^{2}G_L^{\mathrm a},
\qquad
\Xi_L=\mathbb{E}[g_{\mathrm v,L}g_{\mathrm a,L}^{\top}].
\end{aligned}
\label{eq:joint-vs-fusion-factors}
\end{equation}
Under the diagonal approximation used for scoring, the retained difference
has the explicit form
\begin{equation}
\operatorname{diag}
\left(
G_L^{\mathrm{joint}}-G_L^{\mathrm{fusion}}
\right)
=
2\lambda_{\mathrm v}\lambda_{\mathrm a}
\mathbb E[g_{\mathrm v,L}\odot g_{\mathrm a,L}].
\label{eq:joint-diagonal-term}
\end{equation}
Thus, post-hoc Fusion uses both marginal objectives but omits their
coordinatewise alignment before the outer product. Our saliency uses
$G_L=G_L^{\mathrm{joint}}$, with loss weights and normalizers fixed to
their training-time values. This is a gradient-assisted PTQ step: it requires
the original co-training targets and a backward pass, but it does not update
the pretrained weights.

Let $\epsilon_L^{(b)}=Q_b(W_L)-W_L$. A Kronecker-factored
empirical-Fisher approximation~\cite{martens2015optimizing} gives the bit-specific
distortion
\begin{equation}
D_L(b)
=
\frac{1}{2}
\operatorname{tr}\!\left[
G_L\epsilon_L^{(b)}
\Sigma_L(\epsilon_L^{(b)})^\top
\right].
\label{eq:closed-loop-distortion}
\end{equation}
The benefit of upgrading layer $L$ from $b_{\mathrm{lo}}$ to
$b_{\mathrm{hi}}$ is
\begin{equation}
B_L
=
D_L(b_{\mathrm{lo}})
-
D_L(b_{\mathrm{hi}}).
\end{equation}
Layerwise mixed precision is selected by
\begin{equation}
\max_{z_L\in\{0,1\}}
\sum_L z_LB_L,
\qquad
\mathrm{s.t.}\quad
\sum_L z_Lc_L\leq\mathcal B.
\label{eq:bit-allocation}
\end{equation}
The implementation uses a count budget over the candidate Linears:
$c_L=1$ and
$\mathcal B=\lfloor0.2|\mathcal L|\rfloor$. It therefore upgrades the
top 20\% of candidate Linears by $B_L$; the budget is not weighted by the
number of parameters in a Linear. Under diagonal factors, each scalar
contributes
\begin{equation}
s_L(i,j)
=
\frac{1}{2}(G_L)_{ii}(\Sigma_L)_{jj}
\left[
(\epsilon_{L,ij}^{(\mathrm{lo})})^2
-
(\epsilon_{L,ij}^{(\mathrm{hi})})^2
\right].
\label{eq:element-benefit}
\end{equation}
Element-, column-, and layer-level scores are obtained by summing
$s_L(i,j)$ over the scalars contained in the corresponding allocation unit.
In particular,
$C_L(j)=\sum_i s_L(i,j)$ is the column score.

Finite calibration makes fine-grained rankings unstable when adjacent scores
are close. We therefore use layer totals for the precision assignment and
retain column totals only as an ordering heuristic inside GPTQ; element-level
scores are not allocation units. The corresponding top-budget ranking bound
and the distinction between full-ranking recovery and selection stability
are given in Appendix B. GPTQ compensation remains governed by $\Sigma_L$.
The common Hadamard rotation is fused into the weights, whereas smoothing
statistics and diagonal scales may differ by context.

\subsection{Fixed-Intervention Real-Rollout Auditing}
\label{subsec:validity}

Diffusion PTQ is known to require timestep-aware calibration because its
activation distributions vary along the denoising trajectory.
In a WAM, however, the mismatch is
stronger: actions produced by earlier calls change later observations.
A sensitivity profile built from synthetic or open-loop inputs can
therefore measure the right local discrepancy on the wrong states.

Let $q=\{q_t\}_{t=1}^{T}$ be a deployed precision schedule and let $q_0$
denote the unprotected baseline, in which every target module uses the low
precision. Here $t$ indexes the inner denoising step; the outer control-call
index is suppressed. For a recorded FP16 rollout snapshot $x_t$---including
its observation history, chunk position, and persistent cache---define
\begin{equation}
\ell_t(x_t;a)
=
\left\|f_a(x_t)-f_{\mathrm{fp}}(x_t)\right\|_2^2 .
\label{eq:profile-discrepancy}
\end{equation}
The following profiles share this discrepancy but differ in the state
distribution and intervention:
\begin{equation}
\begin{aligned}
S^{\mathrm{syn}}(t)
&=
\mathbb E_{x_t\sim\mathcal D_t^{\mathrm{syn}}}
    [\ell_t(x_t;q_0)],\\
S^{\mathrm{obs}}(t;q)
&=
\mathbb E_{x_t\sim\mathcal D_t^{q}}
    [\ell_t(x_t;q_t)],\\
S_{\mathrm{ref}}^{\mathrm{replay}}(t)
&=
\mathbb E_{x_t\sim\mathcal D_t^{\mathrm{ref}}}
    [\ell_t(x_t;q_0)].
\end{aligned}
\label{eq:three-profiles-final}
\end{equation}
$S^{\mathrm{syn}}$ omits reachable rollout histories; $S^{\mathrm{obs}}$
uses real states but is self-masked by the active protection schedule: a
protected step may appear insensitive because $q_t$ is already active. The
fixed-intervention profile instead replays every recorded state under the
same unprotected intervention $q_0$. We restore the complete immutable FP16
snapshot before each branch, deep-copy persistent state, and match stochastic
seeds between $f_{q_0}$ and $f_{\mathrm{fp}}$. Since the reference
trajectories are generated by the full-precision model, we write
$\mathcal D_t^{\mathrm{ref}}=\mathcal D_t^{\mathrm{fp}}$ rather than calling
it the quantized deployment distribution. A distribution-shift diagnostic
can compare this profile with replay on all-low-bit rollout states through
\begin{equation}
\Delta_{\mathrm{prof}}(t)
=
\left|
S_{\mathcal D^{\mathrm{fp}}}^{\mathrm{replay}}(t)
-
S_{\mathcal D^{q_0}}^{\mathrm{replay}}(t)
\right|.
\label{eq:profile-shift-final}
\end{equation}

This remains a one-call diagnostic. If $s_j$ is the outer closed-loop
state and $\epsilon_j$ the local model error, then to first order
\begin{equation}
\delta s_{j+1}=A_j\delta s_j+B_j\epsilon_j,
\label{eq:closed-loop-localization}
\end{equation}
so downstream task impact also depends on products of transition Jacobians,
not on $\ell_t$ alone. We therefore use profile values to propose a
controlled schedule repair, never as estimates of marginal task gain.

We use the profile only to \emph{audit} an existing $K$-step schedule. Let
$\mathcal T_q$ be its protected set and define
\begin{equation}
\mathcal T_{\mathrm{replay}}
=
\operatorname{TopK}_{t\in[T]}
S_{\mathrm{ref}}^{\mathrm{replay}}(t),
\qquad
|\mathcal T_{\mathrm{replay}}|=|\mathcal T_q|=K.
\label{eq:cf-swap}
\end{equation}
The repaired schedule replaces $\mathcal T_q$ by
$\mathcal T_{\mathrm{replay}}$ without changing the precision levels or
their counts. For each benchmark, the profile is estimated from 32 FP16
closed-loop rollouts that are trajectory-disjoint from the 32-trajectory PTQ
calibration set. The resulting benchmark-level schedule is evaluated on a
separate schedule-validation set and frozen before final testing. Validation
and test trajectories, including their initial-state seeds, are disjoint from
the two 32-rollout sets; task identities may overlap because this is
benchmark-specific calibration rather than held-out-task transfer.
Profile magnitudes are not interpreted as marginal closed-loop gains.
Appendix C gives the complete data flow, and Appendix D specifies snapshot
replay and schedule freezing.

\begin{table}[t]
\centering
\caption{
RoboTwin 2.0 and LIBERO benchmark quantization results on Fast-WAM.
}
\label{tab:fastwam_results}
\renewcommand{\arraystretch}{1.12}
\setlength{\tabcolsep}{5pt}

\resizebox{\textwidth}{!}{%
\begin{tabular}{
    l
    c
    ccc
    c
    |ccccc
    c
    c
}
    \toprule
    \raisebox{-1.7ex}[0pt][0pt]{\textbf{Method}}
    & \raisebox{-1.7ex}[0pt][0pt]{\textbf{Precision}}
    & \multicolumn{3}{c}{\textbf{RoboTwin 2.0}}
    & \raisebox{-1.5ex}[0pt][0pt]{\textbf{Speedup} $\uparrow$}
    & \multicolumn{5}{c}{\textbf{LIBERO}}
    & \raisebox{-1.5ex}[0pt][0pt]{\textbf{Speedup} $\uparrow$}
    & \raisebox{-1.5ex}[0pt][0pt]{\textbf{Mem. (GB)} $\downarrow$} \\
    \cmidrule(lr){3-5}
    \cmidrule(lr){7-11}

    &
    & \textbf{Clean} $\uparrow$
    & \textbf{Random} $\uparrow$
    & \textbf{Average} $\uparrow$
    &
    & \textbf{Goal} $\uparrow$
    & \textbf{Spatial} $\uparrow$
    & \textbf{Object} $\uparrow$
    & \textbf{Long} $\uparrow$
    & \textbf{Average} $\uparrow$
    &
    & \\
    \midrule

    Full Precision
    & FP16
    & \ms{91.9}{0.2}
    & \ms{91.8}{0.6}
    & \ms{91.9}{0.3}
    & 1.0$\times$
    & \ms{98.2}{0.3}
    & \ms{99.8}{0.2}
    & \ms{97.0}{0.3}
    & \ms{95.2}{0.4}
    & \ms{97.6}{0.2}
    & 1.0$\times$
    & 14.4 \\
    \midrule

    GPTQ
    & W4A16
    & \ms{91.2}{0.3}
    & \ms{90.6}{0.6}
    & \ms{90.9}{0.4}
    & 1.2$\times$
    & \ms{95.6}{0.5}
    & \ms{97.5}{0.4}
    & \ms{95.8}{0.5}
    & \ms{95.1}{0.6}
    & \ms{96.0}{0.4}
    & 1.3$\times$
    & 5.5 \\

    SmoothQuant
    & W8A8
    & \ms{91.6}{0.3}
    & \ms{91.0}{0.4}
    & \ms{91.3}{0.3}
    & 1.4$\times$
    & \ms{96.6}{0.4}
    & \ms{97.9}{0.3}
    & \ms{96.3}{0.4}
    & \ms{94.8}{0.5}
    & \ms{96.4}{0.3}
    & 1.5$\times$
    & 7.2 \\
    \midrule

    SVDQuant
    & W4A4
    & \ms{63.8}{0.7}
    & \ms{58.4}{1.0}
    & \ms{61.1}{0.8}
    & 1.6$\times$
    & \ms{73.0}{0.8}
    & \ms{75.1}{0.9}
    & \ms{74.4}{0.9}
    & \ms{72.3}{1.0}
    & \ms{73.7}{0.7}
    & 1.7$\times$
    & 3.6 \\

    $\textbf{SVDQuant}^*$
    & W4A4
    & \ms{66.3}{0.7}
    & \ms{65.5}{0.8}
    & \ms{65.9}{0.7}
    & 1.5$\times$
    & \ms{75.5}{0.7}
    & \ms{76.6}{0.8}
    & \ms{75.7}{0.8}
    & \ms{73.2}{0.9}
    & \ms{75.3}{0.6}
    & 1.5$\times$
    & 4.2 \\
    \midrule

    Atom
    & W4A4
    & \ms{71.5}{0.7}
    & \ms{71.9}{0.8}
    & \ms{71.7}{0.6}
    & 1.6$\times$
    & \ms{75.5}{0.7}
    & \ms{77.6}{0.7}
    & \ms{76.3}{0.8}
    & \ms{75.2}{0.9}
    & \ms{76.2}{0.7}
    & 1.6$\times$
    & 3.8 \\

    $\textbf{Atom}^*$
    & W4A4
    & \ms{78.0}{0.6}
    & \ms{76.4}{0.8}
    & \ms{77.2}{0.6}
    & 1.5$\times$
    & \ms{81.7}{0.6}
    & \ms{84.4}{0.7}
    & \ms{82.0}{0.7}
    & \ms{80.1}{0.8}
    & \ms{82.1}{0.5}
    & 1.6$\times$
    & 4.2 \\
    \midrule

    \rowcolor{gray!15}
    \textbf{QuantWAMs}
    & \textbf{W4A4}
    & \ms{\mathbf{91.8}}{\mathbf{0.2}}
    & \ms{\mathbf{91.6}}{\mathbf{0.5}}
    & \ms{\mathbf{91.7}}{\mathbf{0.3}}
    & 1.4$\times$
    & \ms{\mathbf{98.0}}{\mathbf{0.3}}
    & \ms{\mathbf{99.6}}{\mathbf{0.2}}
    & \ms{\mathbf{96.8}}{\mathbf{0.3}}
    & \ms{\mathbf{95.0}}{\mathbf{0.4}}
    & \ms{\mathbf{97.4}}{\mathbf{0.2}}
    & 1.6$\times$
    & 4.2 \\
    \bottomrule
\end{tabular}%
}

\end{table}

\begin{table}[t]
\centering
\caption{
RoboTwin 2.0 and LIBERO benchmark quantization results on LingBot-VA.
\textbf{Since the officially released LingBot-VA weights support only
LIBERO-Long, results on the other LIBERO suites are not reported.}
}
\label{tab:lingbotva_results}
\renewcommand{\arraystretch}{1.12}
\setlength{\tabcolsep}{5pt}

\resizebox{\textwidth}{!}{%
\begin{tabular}{
    l
    c
    ccc
    c
    |ccccc
    c
    c
}
    \toprule
    \raisebox{-1.7ex}[0pt][0pt]{\textbf{Method}}
    & \raisebox{-1.7ex}[0pt][0pt]{\textbf{Precision}}
    & \multicolumn{3}{c}{\textbf{RoboTwin 2.0}}
    & \raisebox{-1.5ex}[0pt][0pt]{\textbf{Speedup} $\uparrow$}
    & \multicolumn{5}{c}{\textbf{LIBERO}}
    & \raisebox{-1.5ex}[0pt][0pt]{\textbf{Speedup} $\uparrow$}
    & \raisebox{-1.5ex}[0pt][0pt]{\textbf{Mem. (GB)} $\downarrow$} \\
    \cmidrule(lr){3-5}
    \cmidrule(lr){7-11}

    &
    & \textbf{Clean} $\uparrow$
    & \textbf{Random} $\uparrow$
    & \textbf{Average} $\uparrow$
    &
    & \textbf{Goal} $\uparrow$
    & \textbf{Spatial} $\uparrow$
    & \textbf{Object} $\uparrow$
    & \textbf{Long} $\uparrow$
    & \textbf{Average} $\uparrow$
    &
    & \\
    \midrule

    Full Precision
    & FP16
    & \ms{92.9}{0.3}
    & \ms{91.6}{0.5}
    & \ms{92.3}{0.3}
    & 1.0$\times$
    & N/A
    & N/A
    & N/A
    & \ms{98.5}{0.2}
    & \ms{98.5}{0.2}
    & 1.0$\times$
    & 13.5 \\
    \midrule

    GPTQ
    & W4A16
    & \ms{90.9}{0.4}
    & \ms{90.3}{0.6}
    & \ms{90.6}{0.4}
    & 1.3$\times$
    & N/A
    & N/A
    & N/A
    & \ms{97.0}{0.3}
    & \ms{97.0}{0.3}
    & 1.4$\times$
    & 5.6 \\

    SmoothQuant
    & W8A8
    & \ms{91.5}{0.3}
    & \ms{90.9}{0.5}
    & \ms{91.2}{0.3}
    & 1.5$\times$
    & N/A
    & N/A
    & N/A
    & \ms{97.5}{0.3}
    & \ms{97.5}{0.3}
    & 1.6$\times$
    & 6.8 \\
    \midrule

    SVDQuant
    & W4A4
    & \ms{65.8}{0.8}
    & \ms{64.0}{1.0}
    & \ms{64.9}{0.8}
    & 1.5$\times$
    & N/A
    & N/A
    & N/A
    & \ms{73.8}{0.9}
    & \ms{73.8}{0.9}
    & 1.6$\times$
    & 3.4 \\

    $\textbf{SVDQuant}^*$
    & W4A4
    & \ms{69.9}{0.7}
    & \ms{68.5}{0.9}
    & \ms{69.2}{0.7}
    & 1.4$\times$
    & N/A
    & N/A
    & N/A
    & \ms{79.6}{0.8}
    & \ms{79.6}{0.8}
    & 1.5$\times$
    & 3.9 \\
    \midrule

    Atom
    & W4A4
    & \ms{73.0}{0.7}
    & \ms{72.8}{0.9}
    & \ms{72.9}{0.7}
    & 1.6$\times$
    & N/A
    & N/A
    & N/A
    & \ms{76.9}{0.8}
    & \ms{76.9}{0.8}
    & 1.7$\times$
    & 3.6 \\

    $\textbf{Atom}^*$
    & W4A4
    & \ms{77.0}{0.6}
    & \ms{75.5}{0.8}
    & \ms{76.3}{0.6}
    & 1.4$\times$
    & N/A
    & N/A
    & N/A
    & \ms{81.5}{0.7}
    & \ms{81.5}{0.7}
    & 1.5$\times$
    & 3.9 \\
    \midrule

    \rowcolor{gray!15}
    \textbf{QuantWAMs}
    & \textbf{W4A4}
    & \ms{\mathbf{92.3}}{\mathbf{0.3}}
    & \ms{\mathbf{90.9}}{\mathbf{0.5}}
    & \ms{\mathbf{91.6}}{\mathbf{0.3}}
    & 1.4$\times$
    & \textbf{N/A}
    & \textbf{N/A}
    & \textbf{N/A}
    & \ms{\mathbf{98.0}}{\mathbf{0.3}}
    & \ms{\mathbf{98.0}}{\mathbf{0.3}}
    & 1.6$\times$
    & 3.9 \\
    \bottomrule
\end{tabular}%
}

\end{table}

\section{Experiments}\label{sec:experiments}

\subsection{Experimental Settings}\label{subsec:experimental-settings}
We study two released WAM implementations with different coupling patterns.
Fast-WAM uses a dual-stream Mixture-of-Transformers design in which the
video DiT and action expert interact through shared attention. LingBot-VA
uses the officially released shared-backbone checkpoint, in which video and
action tokens traverse the same DiT weights. We use the official checkpoints
and retain the task definitions and evaluation procedures of their released
repositories.

Simulation experiments use RoboTwin 2.0 and LIBERO. RoboTwin 2.0 contains
more than 50 dual-arm tasks and provides 2,500 clean and 25,000 randomized
demonstrations. LIBERO contains the Spatial, Object, Goal, and Long suites,
with ten tasks per suite. The released LingBot-VA checkpoint provides a
LIBERO-Long model only; the unavailable suites are marked N/A rather than
filled by a separately trained checkpoint.

\subsection{Experimental Details}
\paragraph{Data separation.}
For every architecture--benchmark pair, we separate PTQ fitting, rollout
profiling, schedule validation, and final testing. The PTQ calibration set
$\mathcal D_{\mathrm{cal}}$ contains 32 complete trajectories sampled at
random from the benchmark training split. The same 32 trajectories are used
for activation statistics, shared-group screening and mask fitting,
smoothing, co-training-objective saliency, and GPTQ calibration. Group
selection and mask fitting therefore do not use hidden selection and refit
subsets; trajectory-level bootstrap resampling is used only to assess the
stability of the decision.

The profile set $\mathcal D_{\mathrm{prof}}$ contains another 32 closed-loop
rollouts generated by the FP16 model. It is disjoint from
$\mathcal D_{\mathrm{cal}}$ and is used only to construct the
fixed-intervention replay profile. The profile proposes one
benchmark-level precision schedule, which is checked on a separate
full-rollout validation set $\mathcal D_{\mathrm{val}}$ and then frozen.
For each protocol seed, the replay profile on
$\mathcal D_{\mathrm{prof}}^{(r)}$ produces a single Top-$K$ candidate.
$\mathcal D_{\mathrm{val}}^{(r)}$ is used only for a pre-specified
accept--reject comparison against the incumbent schedule. The accepted
schedule is then frozen before evaluation on
$\mathcal D_{\mathrm{test}}^{(r)}$; no schedule search is performed on the
test set. We therefore evaluate a proposal-and-validation procedure rather
than an oracle timestep search.

The four roles are trajectory-disjoint, including initial-state seeds;
task identities may recur across roles. Thus the protocol is
benchmark-specific and in-distribution, but it does not use final-test
trajectories or outcomes for PTQ fitting or schedule selection. Simulation
and real-robot experiments follow the same separation.

Simulation results use end-to-end protocol seeds
$r\in\{42,43,44\}$. For each $r$, we independently draw
$\mathcal D_{\mathrm{cal}}^{(r)}$, $\mathcal D_{\mathrm{prof}}^{(r)}$, and
$\mathcal D_{\mathrm{val}}^{(r)}$, rerun PTQ fitting and schedule selection,
and pair all compared methods on identical final-test initial states.
Across the three seeds, each method is evaluated on 15,000 RoboTwin 2.0
episodes, 6,000 episodes across the four Fast-WAM LIBERO suites, and 1,500
episodes on LingBot-VA LIBERO-Long; each LIBERO-Long ablation configuration
also uses 1,500 episodes.

\begin{figure}
\centering
\includegraphics[
    width=\linewidth,
]{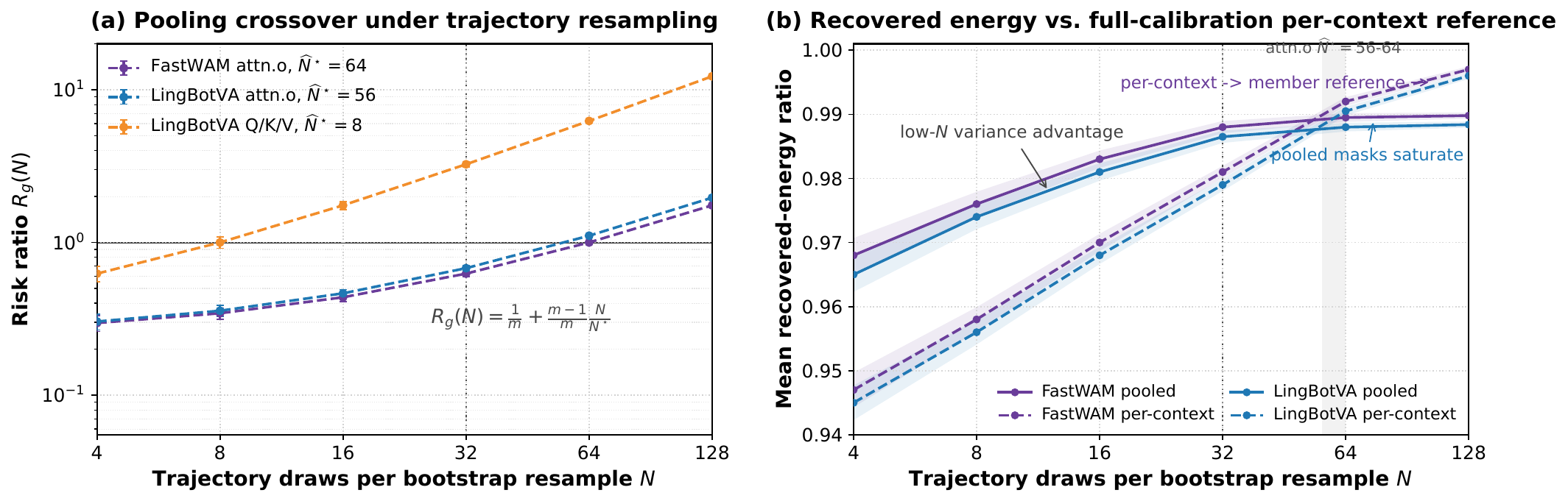}
\caption{\label{fig:pooling_resampling_schematic}
Pooling lowers finite-sample risk and improves recovered Top-$K$ energy
under trajectory-level bootstrap resampling.
}
\vspace{-1em}
\end{figure}

\paragraph{Quantization configuration.}
The default path uses W4A4. Within each admissible activation group, every
member uses the same Hadamard rotation and may use its own diagonal smoothing
scale; statistics are collected after both transforms. The top 2\% input
channels use the BF16 exception path. Weight precision is upgraded to W8A8
for the top 20\% of candidate Linears by count. Fast-WAM protects one of ten
action-denoising steps at A8. LingBot-VA protects two of twenty video steps
and six of fifty action steps at A8. These schedules are fixed per benchmark,
not per task or evaluation trajectory.

The W/A entries under ``Avg. Bits'' describe this nominal execution
schedule. In particular, 4.8 weight bits is the count average
$0.8\times4+0.2\times8$ over candidate Linears; it is not a
parameter-weighted model-size estimate. The activation entry also includes
the 2\% BF16 bypass and the protected-step fractions. Indices, scales, and
other metadata are excluded from this descriptor. We report measured
weight-plus-activation memory separately and give the accounting convention
in Appendix E.

\paragraph{Baselines and calibration information.}
All methods use the same checkpoint, task configuration, and
$\mathcal D_{\mathrm{cal}}$ trajectory identifiers. For the matched-budget
controls, SVDQuant$^*$ and Atom$^*$ use the same candidate modules,
count-based W8 budget, activation outlier fraction, and protected-step
counts as QuantWAMs while retaining their respective quantizers. The
asterisk therefore denotes matched nominal precision allocation, not an
identical calibration algorithm. QuantWAMs additionally uses the original
video--action co-training targets for a backward pass and uses
$\mathcal D_{\mathrm{prof}}$ and $\mathcal D_{\mathrm{val}}$ for schedule
auditing; the conventional PTQ baselines do not receive those signals.

\paragraph{Backend and measurement scope.}
We implement hybrid precision on SM120 Blackwell GPUs: candidate W4 layers
use NVFP4 W4A4 kernels, upgraded layers use FP8 W8A8 kernels, and preserved
outlier channels use a BF16 bypass. RoboTwin 2.0 and LIBERO experiments run
on NVIDIA RTX PRO 5000 Blackwell GPUs. ``Mem. (GB)'' is the measured peak
weight-plus-activation memory of the targeted video and action blocks; it
excludes embedding, projection, VAE, and the remainder of the control
pipeline. ``Speedup'' is the latency ratio for those blocks in one model
call. Neither quantity is an end-to-end control-loop or robot-cycle metric.
Appendix C--E records the split, replay, and measurement procedures.

\subsection{Results in RoboTwin 2.0 and LIBERO}\label{subsec:quantwams-results-in-robotwin-and-libero-benchmarks}

We compare with GPTQ, SmoothQuant, Atom, and the diffusion-model method
SVDQuant, and evaluate every method through closed-loop task success.
Tables~\ref{tab:fastwam_results} and~\ref{tab:lingbotva_results} report the
accuracy--resource trade-off on the two WAMs. Under the W4A4-dominant
configuration, QuantWAMs differs from the corresponding FP16 mean by 0.2
and 0.2 percentage points on Fast-WAM for RoboTwin 2.0 and LIBERO, and by
0.7 and 0.5 points on LingBot-VA. Success rates are reported as
mean $\pm$ sample standard deviation in percentage points over the three
paired protocol seeds. These comparisons are not formal non-inferiority or
equivalence tests.

The matched-budget variants SVDQuant$^*$ and Atom$^*$ control for differences in precision allocation.
Their lower success rates show that the same nominal mixed-precision budget
does not by itself reproduce the QuantWAMs result. This control does not,
however, match the additional gradients and closed-loop rollouts used by
QuantWAMs, nor does it isolate any single component.

For the targeted Fast-WAM blocks, the reported peak
weight-plus-activation memory is 29\% of FP16. The measured speedups are
1.4--1.6$\times$ across the targeted video and action blocks. As specified
above, these are block-level measurements rather than end-to-end deployment
metrics.

\begin{figure}[t]
  \centering
  \begin{minipage}[t]{0.48\textwidth}
    \centering
    \includegraphics[width=\linewidth,keepaspectratio]{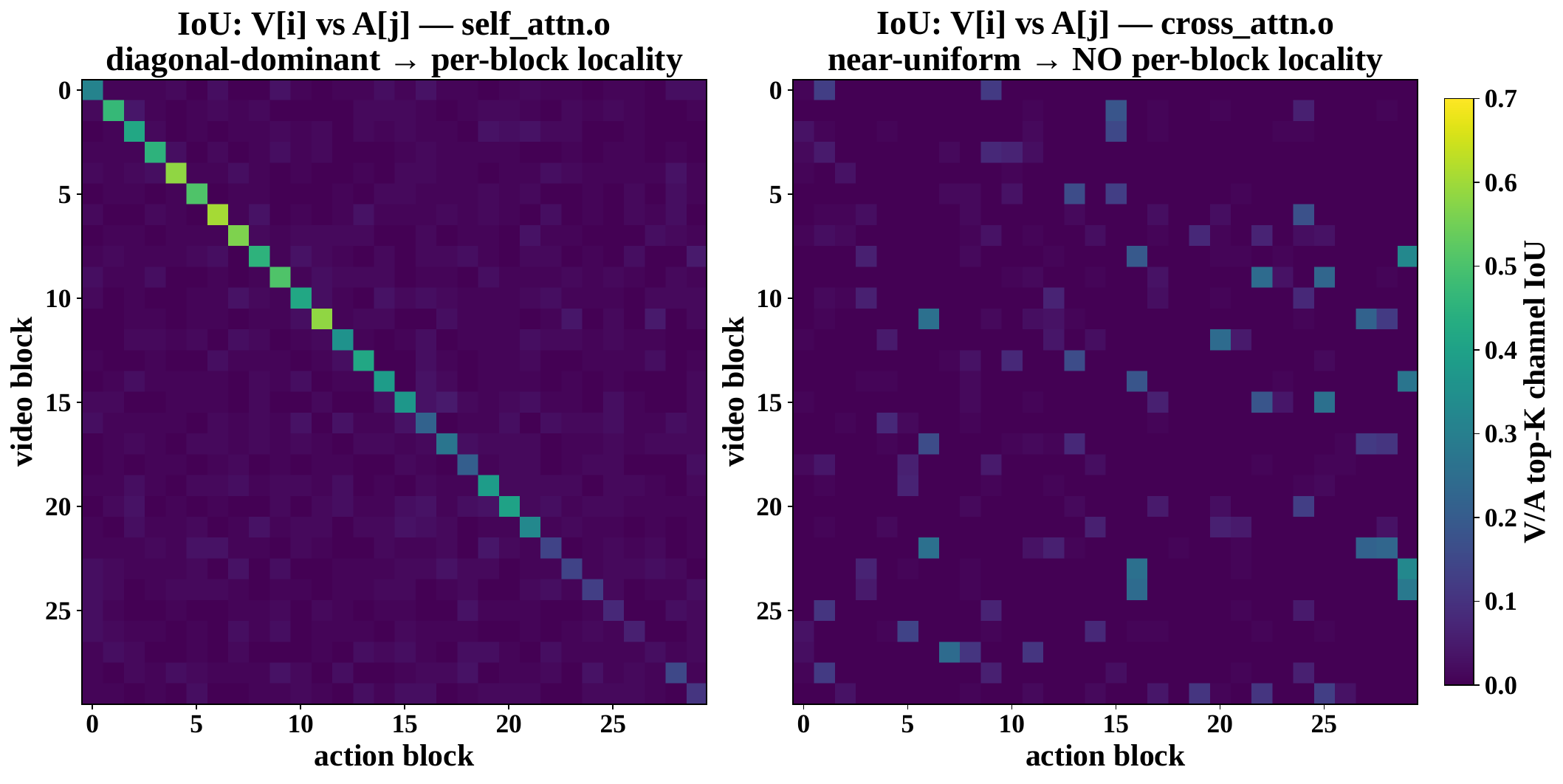}
    \caption{\label{fig:c_raw_locality_cd} V×A IoU matrices. Pairwise top-K channel IoU between video block i and action block j; self.o diagonal-dominant (per-block locality),
  cross.o near-uniform.}
    \vspace{-1.0em}
  \end{minipage}
  \hfill
  \begin{minipage}[t]{0.48\textwidth}
    \centering
    \includegraphics[width=\linewidth,keepaspectratio]{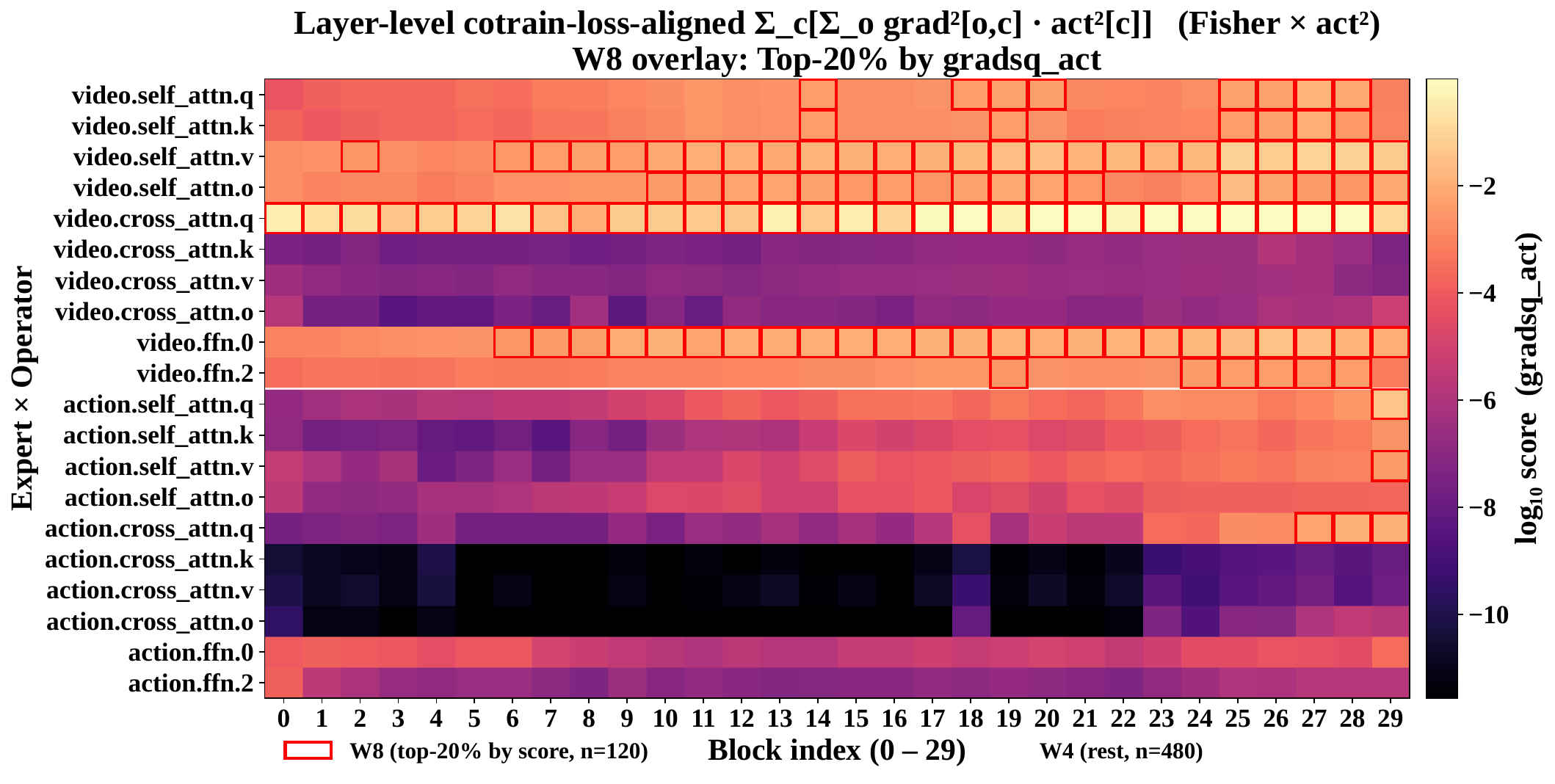}
    \caption{\label{fig:eg-em} Cotrain-loss-aligned per-Linear grad² score across {video, action}×10 operators×30 blocks; red boxes overlay the W8-escalated Linears(top-20\% by score assignment).}
    \vspace{-1.0em}
  \end{minipage}
\end{figure}

\begin{table}[t]
\centering
\footnotesize
\setlength{\tabcolsep}{3pt}
\renewcommand{\arraystretch}{1.10}
\caption{
Fixed-order cumulative ablation on LIBERO-Long under matched precision budgets.
}
\label{tab:cumulative-ladder}
\resizebox{0.65\columnwidth}{!}{%
\begin{tabular}{@{}lccc|cc@{}}
\toprule
& \multicolumn{3}{c|}{Enabled component}
& \multicolumn{2}{c}{LIBERO-Long success (\%) $\uparrow$} \\
\cmidrule(lr){2-4}
\cmidrule(lr){5-6}

Configuration
& \shortstack{Shared\\Basis}
& Joint
& Replay
& Fast-WAM
& LingBot-VA \\
\midrule

Base
& --
& --
& --
& \ms{80.8}{0.7}
& \ms{80.2}{0.9} \\

$+$ Shared-Basis
& $\checkmark$
& --
& --
& \ms{89.1}{0.6}
& \ms{89.6}{0.8} \\

$+$ Joint (Synthetic)
& $\checkmark$
& $\checkmark$
& --
& \ms{90.9}{0.7}
& \ms{91.8}{0.8} \\

\midrule

\textbf{$+$ Replay (Full)}
& $\checkmark$
& $\checkmark$
& $\checkmark$
& \ms{\mathbf{95.0}}{\mathbf{0.4}}
& \ms{\mathbf{98.0}}{\mathbf{0.3}} \\

\bottomrule
\end{tabular}%
}

\end{table}

\subsection{Controlled Ablations}\label{subsec:controlled-ablations}
All ablation experiments in this section were conducted on LIBERO-Long with
matched calibration data and precision budgets. They provide controlled
comparisons of the design choices but do not, by themselves, establish
causal attribution.

\subsubsection{Shared-Basis Pooling: From Recovery to Closed-Loop Success}
Figure~\ref{fig:pooling_resampling_schematic}(a) tests the random-effects prediction
using trajectory-level resamples of the fixed calibration set. For
$M$ siblings,
\[
R(N)=
\frac{\mathrm{MSE}_{\mathrm{pool}}}{\mathrm{MSE}_{\mathrm{per}}}
=
\frac{1}{M}+\frac{M-1}{M}\frac{N}{N^\star},
\]
so the working model predicts a crossover at $R=1$ when $N=N^\star$.
Figure~\ref{fig:pooling_resampling_schematic}(b) compares pooled and
per-Linear masks with a high-$N$ reference using
recovered-energy ratio. Pooling improves recovery at small $N$ and
saturates according to $N_{\mathrm{eff}}$. These resampling curves diagnose
finite-sample behavior on $\mathcal D_{\mathrm{cal}}$; they are not an
independent validation split. Appendix A gives the working-model derivation
and the covariance-aware bootstrap used for the grouping decision.

Figure~\ref{fig:pooling_resampling_schematic} validates the intermediate predictions of
the pooling analysis, whereas Table~\ref{tab:sbc-e2e} tests their
closed-loop consequence. We fix the calibration trajectories, quantizer,
mask size, precision budget, and denoising schedule, varying only mask
grouping. An intermediate within-depth control shares only paired
$\mathrm{attn.o}$ masks inside each block while retaining separate masks
across depths, separating cross-stream sharing from depth collapse in this
controlled comparison.
Screened pooling is compared with per-context masks and an unscreened
global collapse that ignores coordinate validity and the crossover test.
The resulting success rates connect finite-sample mask recovery to
deployment performance.

\subsubsection{Co-Training Saliency and Allocation Granularity}

We evaluate the two axes of our method under matched
scalar-bit budgets. For each allocation granularity, we compute saliency on
the same 32-trajectory calibration set. Trajectory-level bootstrap resamples
of this set are used to assess ranking stability; there is no second fitting
split.
Table~\ref{tab:granularity-e2e} shows that element-level
allocation is unstable and performs poorly in closed loop, whereas
layer-level allocation provides the most reliable precision decision.
Column-level scores are therefore retained only for GPTQ ordering.

We next vary only the saliency objective while fixing the checkpoint,
calibration trajectories, candidate layers, quantizer, and precision
budget. Joint constructs the empirical-Fisher factor from the combined
co-training gradient, whereas post-hoc fusion combines the two
stream-specific factors after the outer product and therefore omits the
cross-stream term in Eq.~\eqref{eq:joint-vs-fusion-factors}. Loss weights
and normalizers are fixed to their training-time values.

At the same precision budget, Joint improves over Fusion by 3.6 and
5.2 percentage points on Fast-WAM and LingBot-VA, respectively.
The variants use identical data, loss weights, and marginal objectives.
Their difference is consistent with retaining the coordinatewise
cross-objective term in Eq.~\eqref{eq:joint-diagonal-term} under our diagonal
approximation, but the closed-loop result alone does not establish a causal
attribution to that term. Joint also outperforms the two single-stream
objectives and the gradient-free activation baseline.

\subsubsection{Fixed-Intervention Rollout Scheduling}
\label{sec:exp-profile-audit}

Fast-WAM protects $K=1$ of its 10 action-denoising steps at A8.
LingBot-VA protects $K=2$ of 20 video-denoising steps and $K=6$ of
50 action-denoising steps at A8. The compared schedules use the same
precision levels, protected-step counts, and fitted quantization parameters;
only the protected indices differ.
Table~\ref{tab:schedule-audit} reports point-estimate differences of
scheduling on Fast-WAM and
LingBot-VA. Since only the protected indices change, this is a
budget-neutral comparison of schedule placement. The replay profiles are
constructed from the 32 FP16 rollouts in $\mathcal D_{\mathrm{prof}}$; the
selected schedule is checked on $\mathcal D_{\mathrm{val}}$ and fixed before
$\mathcal D_{\mathrm{test}}$.

\begin{table*}[t]
\centering

\footnotesize
\setlength{\tabcolsep}{3pt}
\renewcommand{\arraystretch}{1.10}

\begin{minipage}[t]{0.472\textwidth}
\vspace{0pt}
\centering

\begin{minipage}[t][3.2\baselineskip][t]{\linewidth}
\captionof{table}{
Matched-budget structural allocation ablations on LIBERO-Long.
}
\label{tab:structural-ablation}
\label{tab:sbc-e2e}
\label{tab:granularity-e2e}
\end{minipage}

\begin{adjustbox}{max width=\linewidth,center}
\begin{tabular}{@{}lcc@{}}
\toprule
& \multicolumn{2}{c}{LIBERO-Long success (\%) $\uparrow$} \\
\cmidrule(lr){2-3}
Ablation variant
& Fast-WAM
& LingBot-VA \\
\midrule

\multicolumn{3}{@{}l}{\textit{Mask grouping}} \\[-1pt]

\quad Per-context, no pooling
& \ms{82.7}{0.7}
& \ms{82.1}{0.8} \\

\quad Paired-$\mathrm{attn.o}$ pooling
& \ms{84.5}{0.6}
& \ms{81.3}{0.8} \\

\quad Unscreened global
& \ms{86.2}{0.8}
& \ms{85.8}{0.7} \\

\addlinespace[3pt]
\multicolumn{3}{@{}l}{\textit{Weight-allocation granularity}} \\[-1pt]

\quad Element
& \ms{72.1}{0.9}
& \ms{69.8}{1.0} \\

\quad Column
& \ms{82.1}{0.7}
& \ms{81.9}{0.6} \\

\midrule

\shortstack[l]{
    {\scriptsize Screened group + Layer allocation}
}
& \ms{\mathbf{95.0}}{\mathbf{0.4}}
& \ms{\mathbf{98.0}}{\mathbf{0.3}} \\

\bottomrule
\end{tabular}
\end{adjustbox}

\end{minipage}
\hfill
\begin{minipage}[t]{0.49\textwidth}
\vspace{0pt}
\centering

\begin{minipage}[t][3.2\baselineskip][t]{\linewidth}
\captionof{table}{
Matched-budget saliency-source and schedule-selection ablations on
LIBERO-Long.
}
\label{tab:signal-schedule-ablation}
\label{tab:cotrain-saliency}
\label{tab:schedule-audit}
\end{minipage}

\begin{adjustbox}{max width=\linewidth,center}
\begin{tabular}{@{}lcc@{}}
\toprule
& \multicolumn{2}{c}{LIBERO-Long success (\%) $\uparrow$} \\
\cmidrule(lr){2-3}
Ablation variant
& Fast-WAM
& LingBot-VA \\
\midrule

\multicolumn{3}{@{}l}{\textit{Saliency source}} \\[-1pt]

\quad Video-only
& \ms{85.4}{0.7}
& \ms{83.1}{0.8} \\

\quad Action-only
& \ms{87.1}{0.6}
& \ms{86.5}{0.7} \\

\quad Post-hoc fusion
& \ms{91.4}{0.5}
& \ms{92.8}{0.6} \\

\addlinespace[3pt]
\multicolumn{3}{@{}l}{\textit{Schedule selection}} \\[-1pt]

\quad Synthetic Top-$K$
& \ms{90.9}{0.7}
& \ms{91.8}{0.8} \\

\quad Observational Top-$K$
& \ms{93.5}{0.5}
& \ms{93.7}{0.6} \\

\midrule

\shortstack[l]{
    {\scriptsize Joint co-training + Fixed-Intervention}
}
& \ms{\mathbf{95.0}}{\mathbf{0.4}}
& \ms{\mathbf{98.0}}{\mathbf{0.3}} \\

\bottomrule
\end{tabular}
\end{adjustbox}

\end{minipage}

\end{table*}

\subsubsection{Cumulative Component Ladder}
Table~\ref{tab:cumulative-ladder} reports a fixed-order cumulative ablation on LIBERO-Long under matched calibration data and precision budgets.
Base uses per-context masks, post-hoc fusion, and Synthetic Top-$K$.
We then enable Shared-Basis calibration, replace post-hoc fusion with Joint saliency while retaining the synthetic schedule,
and finally replace Synthetic Top-$K$ with Fixed-Intervention Replay to obtain the full method. Within each protocol seed, all configurations use matched calibration draws and identical final-test initial states.

\subsection{Performance on Real-World Robot Tasks}\label{subsec:performance-on-real-world-robot-tasks}

\paragraph{Real-World Environment Setup.}
We used AgiBot G2 as a real-world robot validation platform to perform operational tasks in a real-world environment.
The upper half of the G2 consists of a dual-arm,
7 DoF robotic arm equipped with an end-effector gripper and a three-camera system mounted on the head and wrist.

\paragraph{Tasks and Datasets.}
We evaluate one single-arm task, placing an apple in a basket, and two
dual-arm tasks, stacking three blocks and folding a towel. The corresponding
training collections contain 200, 500, and 700 trajectories. These numbers
describe the training data, not the final evaluation budget. For real-robot
PTQ, $\mathcal D_{\mathrm{cal}}$ is a random 32-trajectory subset of the
training collection. A separate set of 32 FP16 robot rollouts forms
$\mathcal D_{\mathrm{prof}}$; schedule validation and the ten final trials
per task use further disjoint trajectories.

\paragraph{Results.}
In real-world tasks, we used Fast-WAM to evaluate the performance of our quantization algorithm in a real-world pipeline.
We selected ${Atom}^*$, which performed best in the simulation environment, as our baseline for comparison.
Table~\ref{tab:real_world_comparison} presents the results of our real-world experiments.
FP16 succeeds in 19 of 30 trials, QuantWAMs in 17 of 30, and Atom$^*$ in
12 of 30. With ten trials per task, these results establish that the
quantized policy can execute all three tasks, but they do not support an
equivalence or non-inferiority claim relative to FP16. The reported
1.4$\times$ ratio is measured on the targeted WAM blocks and is not an
end-to-end robot-cycle speedup.

\begin{figure}[t]
\label{fig:real-Task}
\centering
\includegraphics[width=\textwidth]{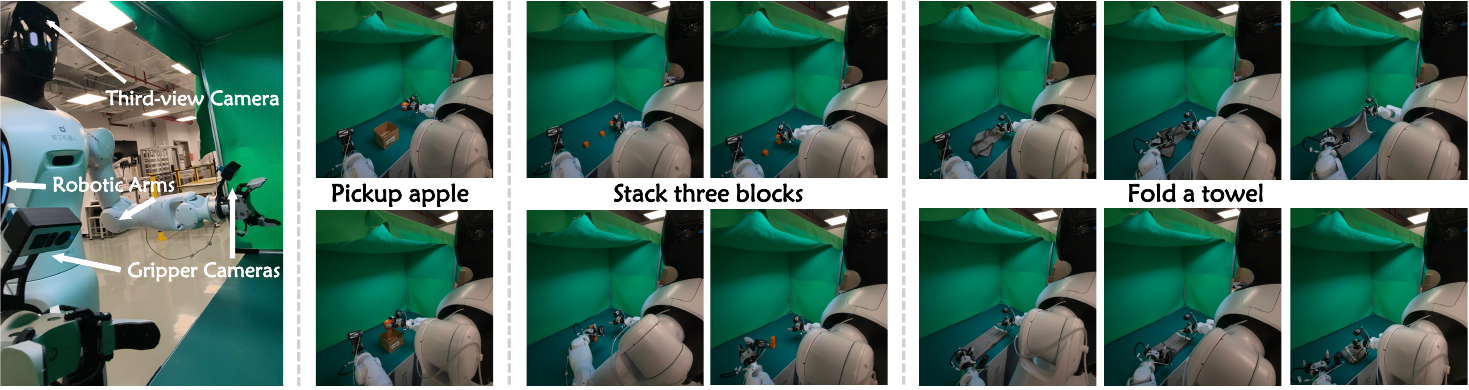} 
\caption{
Real-world system AgiBot G2 and the task examples.
}
\vspace{-1em}
\end{figure}

\begin{table}[t]
    \centering
    \caption{Real-world success rate comparison.
    }
    \label{tab:real_world_comparison}
    \renewcommand{\arraystretch}{1.15}
    \setlength{\tabcolsep}{4pt}

    \resizebox{0.7\linewidth}{!}{%
    \begin{tabular}{
        c|
        c|
        c|
        cc|
        c|
        c
    }
        \toprule

        \raisebox{-1.2ex}[0pt][0pt]{\textbf{Method}}
        &
        \raisebox{-1.2ex}[0pt][0pt]{\textbf{Setting}}
        &
        \multicolumn{1}{c|}{\textbf{One-Arm}}
        &
        \multicolumn{2}{c|}{\textbf{Dual-Arm}}
        &
        \raisebox{-1.2ex}[0pt][0pt]{\textbf{Avg.}}
        &
        \raisebox{-1.2ex}[0pt][0pt]{\textbf{SpeedUp} $\uparrow$}
        \\

        \cmidrule(lr){3-3}
        \cmidrule(lr){4-5}

        &
        &
        \textbf{Pickup apple}
        &
        \textbf{Stack blocks}
        &
        \textbf{Fold a towel}
        &
        &
        \\
        \midrule

        Fast-WAM
        &
        FP16
        &
        8/10
        &
        6/10
        &
        5/10
        &
        63.3\%
        &
        1.0$\times$
        \\

        ${Atom}^*$
        &
        W4A4
        &
        5/10
        &
        4/10
        &
        3/10
        &
        40.0\%
        &
        1.4$\times$
        \\

        \rowcolor{gray!15}
        \textbf{+QuantWAMs}
        &
        \textbf{W4A4}
        &
        8/10
        &
        5/10
        &
        4/10
        &
        56.7\%
        &
        1.4$\times$
        \\

        \bottomrule
    \end{tabular}%
    }
\vspace{-1em}
\end{table}

\section{Conclusion}\label{sec:conclusion}
We presented QuantWAMs, a post-training quantization framework tailored to the closed-loop, multi-pathway nature of World Action Models.
QuantWAMs restricts activation pooling to coordinate-compatible modules,
uses the joint video--action objective for weight saliency, and revises
denoising-step schedules with fixed-intervention replay on FP16 rollout
states. Across the two evaluated WAMs, the W4A4-dominant configurations give
simulation point estimates close to their FP16 counterparts while reducing
the resource use of the targeted blocks.

\section{Limitations}\label{sec:limitations}
QuantWAMs optimizes local surrogates rather than closed-loop performance and requires labels,
one backward pass, FP16 rollouts, and schedule validation.
Results are limited to two WAMs, benchmark-specific calibration,
a Blackwell backend, and block-level efficiency measurements,
without establishing transfer to unseen tasks or end-to-end gains.
The ten-trial robot study demonstrates feasibility but is underpowered
for comparison with FP16.

\bibliography{QuantWAMs}
\bibliographystyle{plainnat}
\end{document}